\documentclass[conference]{IEEEtran}
\IEEEoverridecommandlockouts
\usepackage{graphicx,times,amsmath,amssymb,subfigure,threeparttable,booktabs,multirow,cite}
\usepackage[lined,ruled,commentsnumbered]{algorithm2e}

\makeatletter
\newcommand{\tabincell}[2]{\begin{tabular}{@{}#1@{}}#2\end{tabular}}
\makeatother

\def\BibTeX{{\rm B\kern-.05em{\sc i\kern-.025em b}\kern-.08em
    T\kern-.1667em\lower.7ex\hbox{E}\kern-.125emX}}

\begin{document}

\title{Integrating Informativeness, Representativeness and Diversity in Pool-Based Sequential Active Learning for Regression\\
\thanks{This research was supported by Hubei Technology Innovation Platform under Grant 2019AEA171 and National Natural Science Foundation of China under Grant 61873321.} }

\author{\IEEEauthorblockN{Ziang Liu, Dongrui Wu}
\IEEEauthorblockA{Ministry of Education Key Laboratory of Image Processing and Intelligent Control, \\
School of Artificial Intelligence and Automation, Huazhong University of Science and Technology, Wuhan, China
\\ Email: \{ziangliu,drwu\}@hust.edu.cn}}

\maketitle

\begin{abstract}
In many real-world machine learning applications, unlabeled samples are easy to obtain, but it is expensive and/or time-consuming to label them. Active learning is a common approach for reducing this data labeling effort. It optimally selects the best few samples to label, so that a better machine learning model can be trained from the same number of labeled samples. This paper considers active learning for regression (ALR) problems. Three essential criteria -- informativeness, representativeness, and diversity -- have been proposed for ALR. However, very few approaches in the literature have considered all three of them simultaneously. We propose three new ALR approaches, with different strategies for integrating the three criteria. Extensive experiments on 12 datasets in various domains demonstrated their effectiveness.
\end{abstract}


\section{Introduction}

In many real-world machine learning applications, unlabeled samples are easy to obtain, but it is expensive and/or time-consuming to label them. For example, in emotion estimation from speech signals, speech samples are easy to obtain; however, labeling them requires multiple human experts to listen carefully to score the emotion primitives, e.g., 6-17 experts were used in the VAM corpus \cite{Grimm2008}, and at least 110 experts in IADS-2 \cite{Bradley2007}.

Active learning (AL) \cite{Settles2009} is an effective approach for reducing this data labeling effort. It optimally selects the best few samples to label, so that a better machine learning model can be trained from the same number of labeled samples, compared with random sampling. However, mainly of them considered only classification problems \cite{Abe1998,Cai2014,gal2017deepbayesianAL,Krogh1995,Settles2008,Settles2008b}.

This paper considers only regression problems, i.e., active learning for regression (ALR). There are two scenarios in ALR: \emph{population-based} and \emph{pool-based} \cite{Sugiyama2009}. This paper considers the latter, where a pool of unlabeled samples is given, and we need to optimally select few of them to label, so that a regression model trained from them can be used to label other samples. The samples are selected iteratively, and only one sample is selected in each iteration. This situation is common in off-line machine learning applications \cite{drwuSAL2019}.

Three essential criteria for ALR have been proposed in our previous work \cite{drwuSAL2019}:
\begin{enumerate}
    \item \emph{Informativeness}, which means that the selected samples should contain rich information, so labeling them would contribute more to the regression function.
    \item \emph{Representativeness}, which means the sample density around each selected sample should be high, so that each selected sample can represent more neighbouring samples and less likely to be an outlier.
    \item \emph{Diversity}, which means the selected samples should scatter across the entire input space, instead of concentrating on a small region of it.
\end{enumerate}
We \cite{drwuSAL2019} have also demonstrated empirically that each individual criterion contributes to the overall ALR performance.

Traditional ALR approaches, e.g., query-by-committee (QBC) \cite{Burbidge2007,Cohn1996,Krogh1995,RayChaudhuri1995,Settles2009,Demir2014} and expected model change maximization (EMCM) \cite{Cai2013,Cai2017}, consider only the informativeness. We \cite{drwuSAL2019} proposed a representativeness-diversity (RD) approach to consider simultaneously the representativeness and the diversity. RD can also be integrated with QBC or EMCM to consider the informativeness, representativeness and diversity simultaneously. At the same time, \cite{drwuiGS2019} proposed an improve greedy sampling (iGS) approach to consider both the informativeness and diversity, and demonstrated that iGS outperformed QBC and EMCM.

However, no one has investigated if integrating RD and iGS can improve the performance of both, and also outperform RD-QBC and RD-EMCM. This paper fills this gap by proposing three new strategies to integrate RD and iGS. Our main contributions are:
\begin{enumerate}
\item We consider ALR in two separate processes: the initialization process and the iteration process. The initialization process selects few samples from the unlabeled sample pool to label and builds the initial regression model. The sample selection is completely unsupervised, thus only the representativeness and the diversity can be considered. An iteration process selects more samples to label, using the previously built regression model, and updates the regression model. The iteration process is supervised, and hence all three criteria can be considered.
\item We propose three new strategies to integrate RD and iGS, and hence the informativeness, representativeness and diversity are considered simultaneously.
\item We performed experiments on various datasets from diverse application domains to verify the effectiveness of our proposed ALR approaches. Particularly, two of them performed much better than the state-of-the-art ALR approaches.
\end{enumerate}

The remainder of this paper is organized as follows: Section~\ref{sec:existing strategies} introduces five state-of-the-art ALR approaches, which will be compared with our proposed ALR approaches. Section~\ref{sec:proposed_approaches} proposes three new approaches for integrating the informativeness, representativeness, and diversity in ALR. Section~\ref{sect:experiments} compares the performances of our proposed ALR approaches with another six approaches. Finally, Section~\ref{sect:conclusions} draws conclusions.

\section{Existing ALR Strategies} \label{sec:existing strategies}

This section briefly introduces five state-of-the-art pool-based sequential ALR approaches. Each considers one or more of the three criteria. We assume the pool consists of $N$ unlabeled samples $\{\mathbf{x}_i\}^N_{i=1}$, the initialization process selects $M_0$ samples to label, and each iteration processes selects one more sample to label.

QBC \cite{RayChaudhuri1995}, proposed in 1995, is probably the first ALR approach. It uses random sampling in the initialization process. In each sequential iteration, it first bootstraps the labeled samples and builds a committee of regression models. Then, for each unlabeled sample in the pool, it predicts its labels by the committee and computes the variance. Finally, it selects the sample with maximum variance to label. Clearly, QBC considers the informativeness only.

GSx (Greedy Sampling in the Input Space) \cite{Yu2010,drwuiGS2019}, first proposed in 2010, is a passive sampling approach that considers the diversity only. It selects the sample closest to the centroid of the whole sample pool as the first sample to label, and then in each sequential iteration a sample furthest away from all existing selected samples to label. The selected samples can be labeled at once after all of them are determined, because the selection process does not need the label information.

More specifically, let $\{\mathbf{x}_m\}^M_{m=1}$ be the already selected samples. For each of the $N-M$ unlabeled samples, GSx computes its closest distance to the $M$ labeled samples, and then selects the sample $\mathbf{x}^*$ with the maximum distance to label:
\begin{align}
d^x_n &= \min_{m}||\mathbf{x}_n-\mathbf{x}_m||, \quad m=1,...,M; n=M+1,...,N \label{eq:GSx_dx} \\
\mathbf{x}^* &= \arg\max_{\mathbf{x}_n}\ d^x_n,
\end{align}

EMCM \cite{Cai2013}, proposed in 2013, also considers the informativeness only. In the initialization process, it also randomly selects $M_0$ samples to label, and train a regression model from them. In each sequential iteration, it selects the sample that may change the model parameters the most to label. More specifically, it uses all $M$ already labeled samples to train a regression model, whose prediction for the $n$th sample $\mathbf{x}_n$ is $\hat{y}_n$. It also uses bootstrap on the $M$ already labeled samples to construct $P$ linear regression models. Let the $p$th model's prediction for the $n$th sample $\mathbf{x}_n$ be $y_n^p$. Then, for each of the $N-M$ unlabeled samples, it computes
\begin{align}
g(\mathbf{x}_n)=\frac{1}{P}\sum_{p=1}^P\left\| (y_n^p-\hat{y}_n)\mathbf{x}_n\right\|, \quad n=M+1,...,N
\end{align}
EMCM then selects the sample with the maximum $g(\mathbf{x}_n)$ to label.

RD-EMCM \cite{drwuSAL2019}, proposed in 2019, integrates RD and EMCM, and is the best-performing ALR in \cite{drwuSAL2019}. Its initialization process uses RD only, i.e., it performs $k$-means ($k=M_0$) clustering of all $N$ samples and selects the one closest to each cluster centroid to label. In each sequential iteration, it performs $k$-means clustering ($k=M+1$, where $M$ is the number of labeled samples) on the $N$ samples, finds the largest cluster that does not contain an already labeled sample, and performs EMCM in that cluster to select one sample to label.

iGS \cite{drwuiGS2019}, proposed in 2019, improves GSx by using greedy sampling in both the feature space and the label space. GSx is used in the initialization process. In each sequentially iteration, for all unlabeled samples, it first uses (\ref{eq:GSx_dx}) to compute $d^x_n$ in the input space, and then computes $d^y_n$ in the label space:
\begin{align}
d^y_n &= \min_{m}||\hat{y}_n-y_m||,\quad m=1,...,M; n=M+1,...,N \label{eq:iGS_dy}
\end{align}
It next computes $d^{xy}_n$ to integrate $d^x_n$ and $d^y_n$:
\begin{align}
d^{xy}_n &= d^x_n\cdot d^y_n,\quad n=M+1,...,N \label{eq:iGS_dxy}
\end{align}
Finally, it selects the sample $\mathbf{x}^*$ with the maximum $d^{xy}_n$ to label:
\begin{align}
\mathbf{x}^* &= \arg\max_{\mathbf{x}_n}\ d^{xy}_n
\end{align}
Our previous work \cite{drwuiGS2019} has shown that iGS outperforms QBC, EMCM and GSx.

\section{Three New Strategies to Integrate RD and iGS} \label{sec:proposed_approaches}

In this section, we propose three new strategies to integrate RD and iGS, so that the informativeness, representativeness and diversity are considered simultaneously in ALR.

\subsection{RD-iGS}

RD-iGS integrates RD and iGS, similar to the way of integrating RD and EMCM in \cite{drwuSAL2019}. Its initialization process uses RD only, i.e., it performs $k$-means ($k=M_0$) clustering of all $N$ samples and selects the one closest to each cluster centroid to label. In each sequential iteration, it performs $k$-means clustering ($k=M+1$, where $M$ is the number of labeled samples) on the $N$ samples, finds the largest cluster that does not contain an already labeled sample, and performs iGS in that cluster to select one sample to label.

Comparing against the three criteria for ALR, RD considers the diversity and the representativeness, and iGS considers the diversity and the informativeness. Thus, RD-iGS considers all three criteria. The pseudo-code of RD-iGS is shown in Algorithm~\ref{alg:RD_iGS}.

\begin{algorithm}[!h] 
\KwIn{A pool of $N$ unlabeled samples, $\{\mathbf{x}_i\}^N_{i=1}$\;
    \hspace*{10mm} $M_{\max}$, the maximum number of samples to label.}
\KwOut{The regression model $f(\mathbf{x})$.}
\tcp{Initialization process}
Perform $k$-means ($k=M_0$) clustering on $\{\mathbf{x}_i\}^N_{i=1}$\;
Select from each cluster one sample closest to its centroid, query for its label, and add it to the training set $T$\;
Construct the initial regression model $f(\mathbf{x})$\;
\tcp{Iteration process}
\For{$M=M_0+1,...,M_{\max}$}{
    Perform $k$-means ($k=M$) clustering on $\{\mathbf{x}_i\}^N_{i=1}$\;
    Identify the largest cluster $C_{\max}$ that does not already contain any labeled sample\;
    Compute $d^{xy}$ in (\ref{eq:iGS_dxy}) for each sample in $C_{\max}$ \;
    Select the sample with the maximum $d^{xy}$, query for its label, and add it to the training set $T$\;
    Update the regression model $f(\mathbf{x})$ using $T$\;}
\caption{The proposed RD-iGS algorithm.} \label{alg:RD_iGS}
\end{algorithm}

\subsection{RDiGS}

In RDiGS, RD is used for initialization and iGS for iteration. It performs unsupervised RD in the initialization process, as in RD-iGS, and performs iGS in the iteration process.

The pseudo-code of RDiGS is shown in Algorithm~\ref{alg:RDiGS}. It also considers all three criteria for ALR.

\begin{algorithm}[!h] 
\KwIn{A pool of $N$ unlabeled samples, $\{\mathbf{x}_i\}^N_{i=1}$\;
    \hspace*{10mm} $M_{\max}$, the maximum number of samples to label.}
\KwOut{The regression model $f(\mathbf{x})$.}
\tcp{Initialization process}
Perform $k$-means ($k=M_0$) clustering on $\{\mathbf{x}_i\}^N_{i=1}$\;
Select from each cluster one sample closest to its centroid, query for its label, and add it to the training set $T$\;
Construct the initial regression model $f(\mathbf{x})$\;
\tcp{Iteration process}
\For{$M=M_0+1,...,M_{\max}$}{
    Compute $d^{xy}$ in (\ref{eq:iGS_dxy}) for each unlabeled sample $\{\mathbf{x}_n\}^N_{n=M}$\;
    Select the sample with the maximum $d^{xy}$, query for its label, and add it to the training set $T$\;
    Update the regression model $f(\mathbf{x})$ using $T$\;}
\caption{The proposed RDiGS algorithm.} \label{alg:RDiGS}
\end{algorithm}

\subsection{RDiGSr} \label{alg:RDiGSr}

RDiGSr performs RD in the initialization process, and iGSr in the iteration process. iGSr first uses (\ref{eq:GSx_dx}) to compute $d^x_n$ in the input space and (\ref{eq:iGS_dy}) to compute $d^y_n$ in the label space. It then uses (\ref{eq:iGS_dxy}) to compute $d^{xy}_n$, just as iGS. Finally, it computes the representativeness $R_n$ for all unlabeled samples:
 \begin{align}
R_n = \sum^{N}_{i=1}||\mathbf{x}_n-\mathbf{x}_i|| \label{eq:iGSr_R}
\end{align}
and selects the one with the maximum $d^{xy}_n/R_n$ to label:
\begin{align}
\mathbf{x}^* = \arg\max_{\mathbf{x}_n}(d^{xy}_n/R_n),\quad n=M+1,...,N \label{eq:iGSr}
\end{align}

The pseudo-code of RDiGSr is shown in Algorithm~\ref{alg:RDiGSr}. It also considers all three criteria for ALR.

\begin{algorithm}[!h] 
\KwIn{A pool of $N$ unlabeled samples, $\{\mathbf{x}_i\}^N_{i=1}$\;
    \hspace*{10mm} $M_{max}$, the maximum number of samples to label.}
\KwOut{The regression model $f(\mathbf{x})$.}
\tcp{Initialization process}
Perform $k$-means ($k=M_0$) clustering on $\{\mathbf{x}_i\}^N_{i=1}$\;
Select from each cluster one sample closest to its centroid, query for its label, and add it to the training set $T$\;
Construct the initial regression model $f(\mathbf{x})$\;
\tcp{Iteration process}
\For{$M=M_0+1,...,M_{max}$}{
    Compute $d^{xy}$ in (\ref{eq:iGS_dxy}) for each unlabeled sample $\{\mathbf{x}_n\}^N_{n=M}$\;
    Compute $R$ in (\ref{eq:iGSr_R}) for each unlabeled sample $\{\mathbf{x}_n\}^N_{n=M}$\;
    Use (\ref{eq:iGSr}) to select one sample, query for its label, and add it to the training set $T$\;
    Update the regression model $f(\mathbf{x})$ using $T$\;}
\caption{The proposed RDiGSr algorithm.} \label{alg:RDiGSr}
\end{algorithm}

\subsection{Summary}

Table~\ref{tab:8algs_3criteria} summarizes how the informativeness, representativeness and diversity are considered in different ALR approaches.

\begin{table}[!h]   \centering
  \caption{Summary of how the informativeness (I), representativeness (R) and diversity (D) are considered in different ALR approaches. RS means random sampling.}
    \begin{tabular}{c|cccc|cccc}
    \toprule
      & \multicolumn{4}{c|}{Initialization} & \multicolumn{4}{c}{Iteration} \\
      & Alg & I & R & D & Alg & I & R & D \\
    \midrule
    QBC & RS & -- & -- & -- & QBC & $\checkmark$ & -- & --\\
    GSx & GSx & -- & -- & $\checkmark$ & GSx &-- & -- &$\checkmark$ \\
    EMCM & RS & -- & -- & -- & EMCM &$\checkmark$ & -- &-- \\
    RD-EMCM & RD & -- & $\checkmark$ &$\checkmark$ & RD-EMCM & $\checkmark$ & $\checkmark$ & $\checkmark$ \\
    iGS & GSx & -- & -- & $\checkmark$& iGS & $\checkmark$ & -- & $\checkmark$ \\
    RD-iGS & RD & -- & $\checkmark$ &$\checkmark$ & RD-iGS & $\checkmark$ & $\checkmark$ & $\checkmark$ \\
    RDiGS & RD &-- & $\checkmark$ & $\checkmark$& iGS & $\checkmark$ & -- & $\checkmark$ \\
    RDiGSr & RD &-- & $\checkmark$ & $\checkmark$ & iGSr & $\checkmark$ & $\checkmark$ & $\checkmark$ \\
    \bottomrule
    \end{tabular}%
  \label{tab:8algs_3criteria}%
\end{table}%

\section{Experiments and Results}\label{sect:experiments}

Extensive experiments are performed in this section to demonstrate the advantages of our proposed three ALR approaches over five state-of-the-art ALR approaches.

\subsection{Datasets} \label{sect:Datasets}

Table~\ref{tab:Datasets} shows the summary of the 12 datasets used in our experiments. They cover a wide variety of application domains. More of them are from the UCI Machine Learning Repository\footnote{http://archive.ics.uci.edu/ml/index.php} and the CMU StatLib Datasets Archive\footnote{http://lib.stat.cmu.edu/datasets/}, which had also been used in many previous ALR experiments \cite{Cai2017,Cai2013,Yu2010,drwuSAL2019,drwuiGS2019}. We also used an affective computing dataset \emph{Vera am Mittag} (VAM; \emph{Vera at Noon} in English) \cite{Grimm2008}), which has been used in many previous studies \cite{Grimm2007b,Grimm2007a,drwuICME2010,drwuInterSpeech2010,drwuMTALR2020}. Only arousal in VAM was used as the regression output.

Two datasets (autoMPG and CPS) contained both numerical and categorical features. For them, we first used one-hot coding to covert the categorical values into numerical values, which increased their dimensionality. We then used principal component analysis to reduce their dimensionality back to their original number of dimensions. For each dataset, we normalized each dimension of the input to mean zero and standard deviation one.

\begin{table}[!h] \centering  \setlength{\tabcolsep}{1mm}
\caption{Summary of the 12 regression datasets.}   \label{tab:Datasets}
\begin{tabular}{c|cccccc}   \hline
Dataset     &\tabincell{c}{No. of\\samples} &\tabincell{c}{No. of\\raw\\features} &\tabincell{c}{No. of\\numerical\\features} &\tabincell{c}{No. of\\categorical\\features} &\tabincell{c}{No. of\\total\\features}    \\ \hline
Concrete-CS    &103            &7                   &7                         &0           &7\\
Yacht          &308            &6                   &6                         &0           &6\\
autoMPG        &392            &7                   &6                         &1           &9\\
NO2            &500            &7                   &7                         &0           &7\\
PM10           &500            &7                   &7                         &0           &7\\
Housing        &506            &13                  &13                        &0          &13 \\
CPS            &534            &10                  &7                         &3          &19\\
VAM-Arousal     &947            &46                  &46                        &0         &46\\
Concrete        &1,030           &8                   &8                         &0         &8\\
Airfoil       &1,503           &5                   &5                         &0        &5\\
Wine-Red      &1,599           &11                  &11                        &0        &11\\
Wine-White    &4,898           &11                  &11                        &0         &11\\ \hline
\end{tabular}
\end{table}

\subsection{Performance Evaluation Process}

For each dataset, we randomly selected $80\%$ samples as the training pool, and the remaining $20\%$ as the test set. We used the mean and the variance of the training samples to normalize the test samples, because in practice the test samples are unknown.

We compared the performances of our proposed RD-iGS, RDiGS and RDiGSr with six approaches: random sampling (RS), QBC, GSx, EMCM, RD-EMCM, and iGS. For each sampling approach, we selected $M\in[d+1, d+21]$ samples, where $d$ is the feature dimensionality after data preprocessing\footnote{A reliable linear regression model requires at least $d+1$ training samples, where $d$ corresponds to the $d$ coefficients of the $d$ features, and 1 corresponds to the bias term.}. We then labeled them and built a ridge regression model\footnote{We used the function \emph{fitrlinear} in MATLAB 2019, which performs better than the function \emph{ridge} used in our previous work \cite{drwuSAL2019,drwuiGS2019}, especially when $M$ is small. The performance improvements of all ALR approaches decreased as the performance of RS improved a lot.} with the L2 regularization coefficient $r=0.01$. We used the root mean squared error (RMSE) and the correlation coefficient (CC) as performance measures. The above process was repeated 100 times to obtain statistically meaningful results.

\subsection{Results} \label{sect:performance}

\begin{figure*}[!h]\centering
\includegraphics[width=\linewidth,clip]{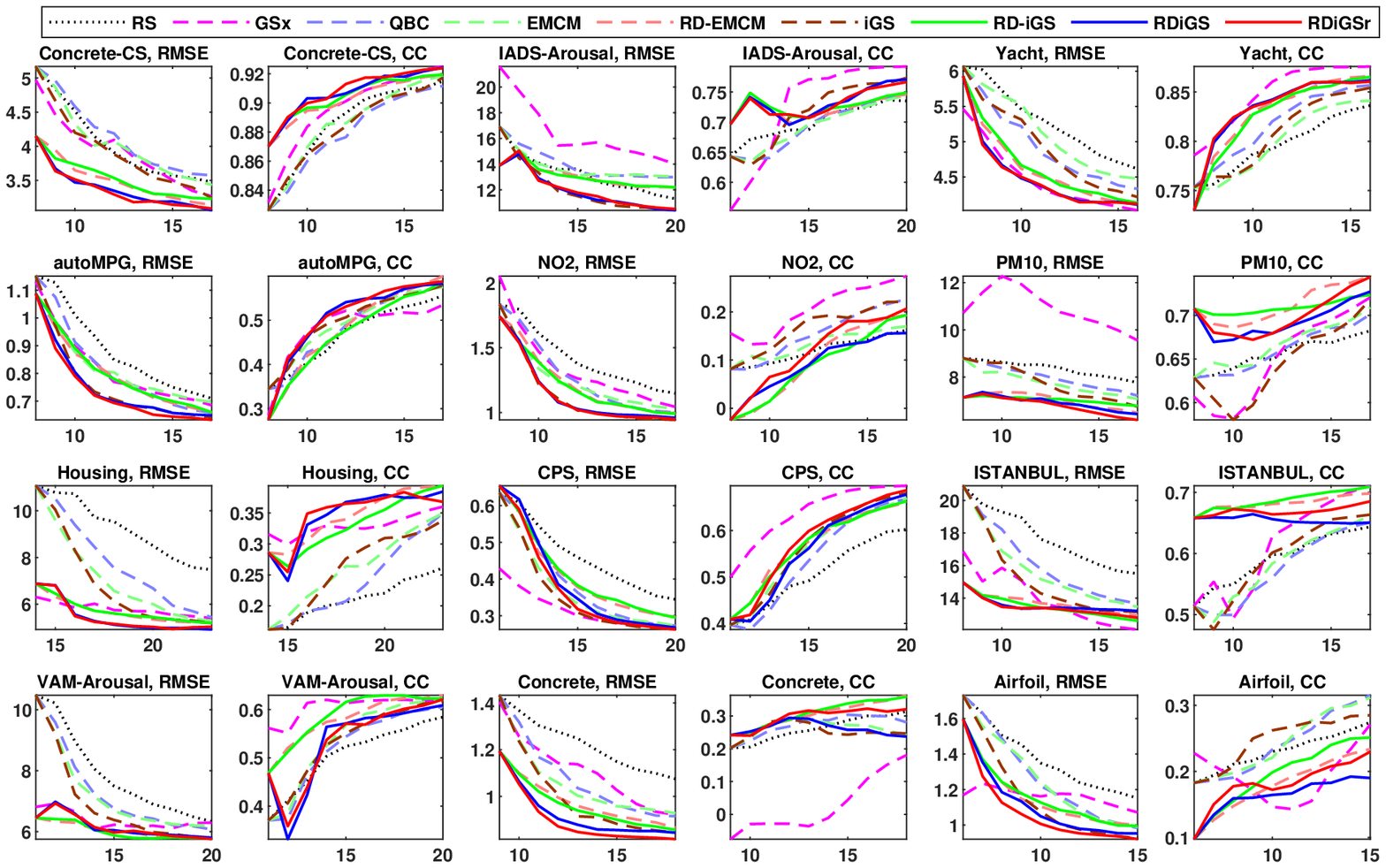}
\caption{Mean RMSEs and mean CCs of the nine sampling approaches on the 12 datasets, averaged over 100 runs. The horizontal axis represents $M$, the number of samples to be labeled.} \label{fig:curve_9algs}
\end{figure*}

Fig.~\ref{fig:curve_9algs} shows the performances of the nine approaches, each averaged across the 100 runs. RDiGSr achieved the smallest RMSE on most datasets and for most $M$.

\begin{figure}[!h]\centering
\includegraphics[width=\linewidth,clip]{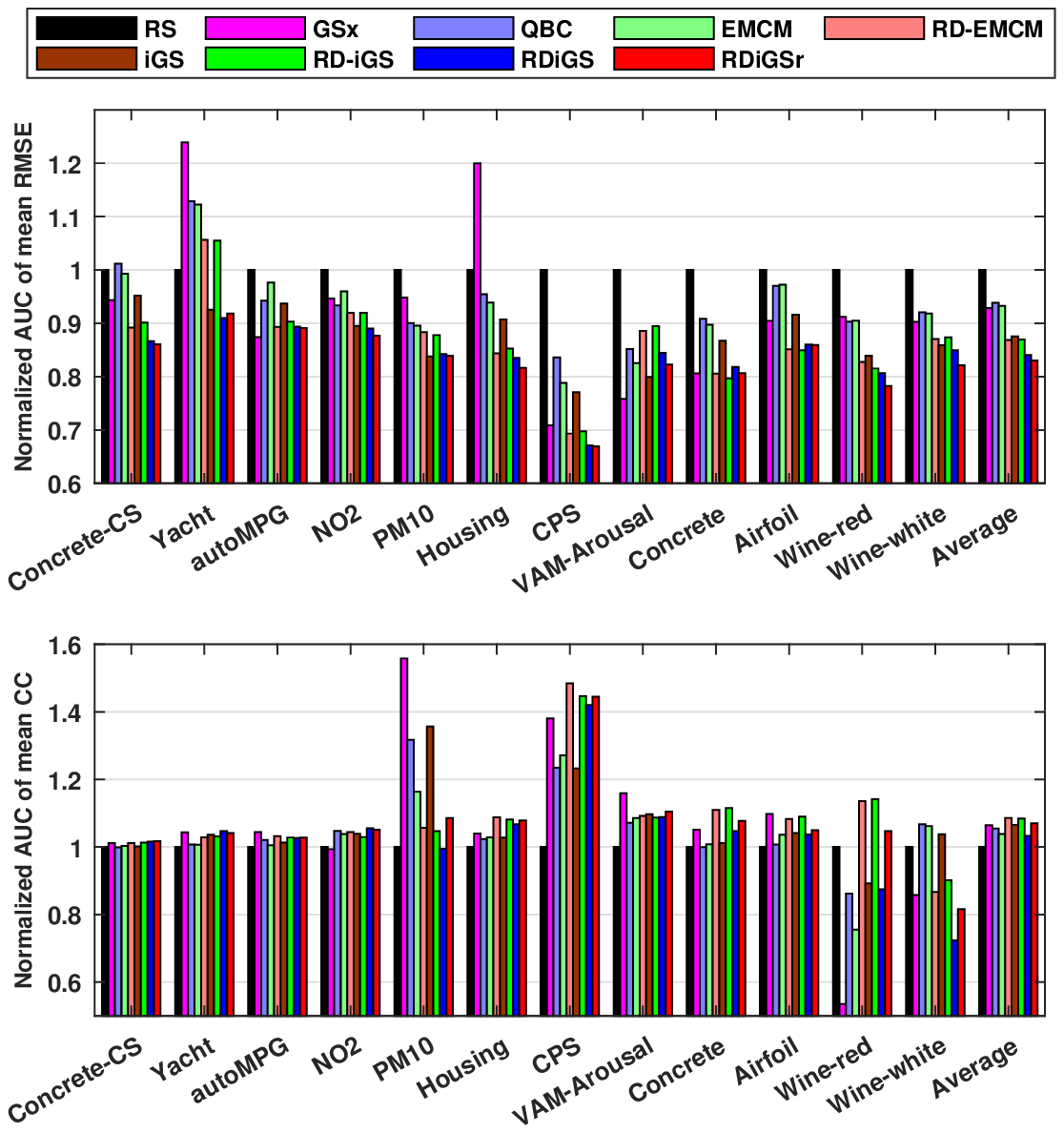}
\caption{Normalized AUCs of the mean RMSEs and the mean CCs on the 12 datasets. Ridge regression ($r=0.01$) was used as the regression model.} \label{fig:bar_9algs_ridge001}
\end{figure}

To compare the performances of different algorithms more directly, we computed the area under the curves (AUCs) of the mean RMSEs and the mean CCs. Because the AUCs from different datasets varied a lot, we normalized them w.r.t. the AUC of RS on each dataset; thus, the AUC of RS was always 1 on each dataset. Fig.~\ref{fig:bar_9algs_ridge001} shows the results, and Table~\ref{tab:imptfRS} shows the improvements of the eight ALR approaches over the RS baseline. On average, all ALR approaches had smaller RMSE and larger CC than RS. The differences between the RMSEs of different ALR approaches were more obvious than the differences between their CCs. On average, our proposed RDiGSr had the smallest RMSE.

\begin{table}[!h] \setlength{\tabcolsep}{0.9mm} \footnotesize   \centering
  \caption{Percentage improvements of the AUCs of the mean RMSEs and the mean CCs over RS. The best performances are marked in bold.}
    \begin{tabular}{cc|cccccccc}     \toprule
      &   & GSx & QBC & EMCM & \multicolumn{1}{c}{\tabincell{c}{RD-\\EMCM}} & iGS & \multicolumn{1}{c}{\tabincell{c}{RD-\\iGS}} & RDiGS & RDiGSr \\
    \midrule
    \multirow{2}[2]{*}{RMSE} & Mean & 7.1  & 6.2  & 6.7  & 13.1  & 12.5  & 13.0  & 15.9  & \textbf{17.0}  \\
      & Var & 61.9  & 35.8  & 41.2  & 70.4  & 47.5  & 69.1  & 75.3  & \textbf{75.4}  \\
    \midrule
    \multirow{2}[2]{*}{CC} & Mean & 6.4  & 5.5  & 3.9  & \textbf{8.6}  & 6.6  & 8.4  & 3.3  & 7.0  \\
      & Var & 39.5  & 11.7  & 11.1  & 39.2  & 15.6  & \textbf{39.9}  & 30.0  & 34.5  \\
    \bottomrule
    \end{tabular}%
  \label{tab:imptfRS}%
\end{table}%

To determine whether the performance improvements of our proposed three ALR approaches over the other two best performing ALR approaches (RD-EMCM and iGS) were statistically significant, we used two-sample one-tailed $t$-test (MATLAB2019 function \emph{ttest2}) with the significance level $\alpha=0.05$ on the average AUCs between each pair of ALR approaches. The results are shown in Table~\ref{tab:ttest2}. Both RDiGS and RDiGSr had significantly smaller RMSEs than RD-EMCM and iGS.

Comparing RDiGSr with RDiGS, RDiGSr had smaller RMSE ($p=0.0690$, close to the threshold $p=0.05$) and larger CC ($p=0.0001$) than RDiGS, which demonstrated the advantages of iGSr (Section~\ref{alg:RDiGSr}) over iGS (Section~\ref{sec:existing strategies}), both of which are sequential sampling approaches used in the iteration process of ALR.

\begin{table}[!h]   \centering
  \caption{$p$-values of the $t$-tests. The alternative hypothesis is that the row-wise ALR approach performed better (had smaller RMSE or larger CC) than the column-wise ALR approach. A number in bold means the alternative hypothesis is accepted.}
    \begin{tabular}{cc|cc}
    \toprule
      & \multicolumn{1}{c|}{} & \multicolumn{1}{c}{RD-EMCM} & \multicolumn{1}{c}{iGS}  \\
      \midrule
    \multicolumn{1}{c}{\multirow{2}[2]{*}{RD-iGS}} & RMSE & .4904 & \textbf{.0000}  \\
      & CC & .5839 & \textbf{.0000} \\
    \midrule
    \multicolumn{1}{c}{\multirow{2}[2]{*}{RDiGS}} & RMSE & \textbf{.0000} & \textbf{.0000}  \\
      & CC & 1.0000 & .6713  \\
    \midrule
    \multicolumn{1}{c}{\multirow{2}[2]{*}{RDiGSr}} & RMSE & \textbf{.0000} & \textbf{.0000}  \\
      & CC & .9707 & \textbf{.0010}  \\
    \bottomrule
    \end{tabular}%
  \label{tab:ttest2}%
\end{table}%

In summary, all of our three proposed ALR approaches, RD-iGS, RDiGS and RDiGSr, performed better than or comparable with the state-of-the-art RD-EMCM and iGS approaches in the literature. Among the three, RDiGSr may be the best.

\subsection{Sensitivity to the Regularization Coefficient}

\begin{figure}[!h]\centering
\includegraphics[width=\linewidth,clip]{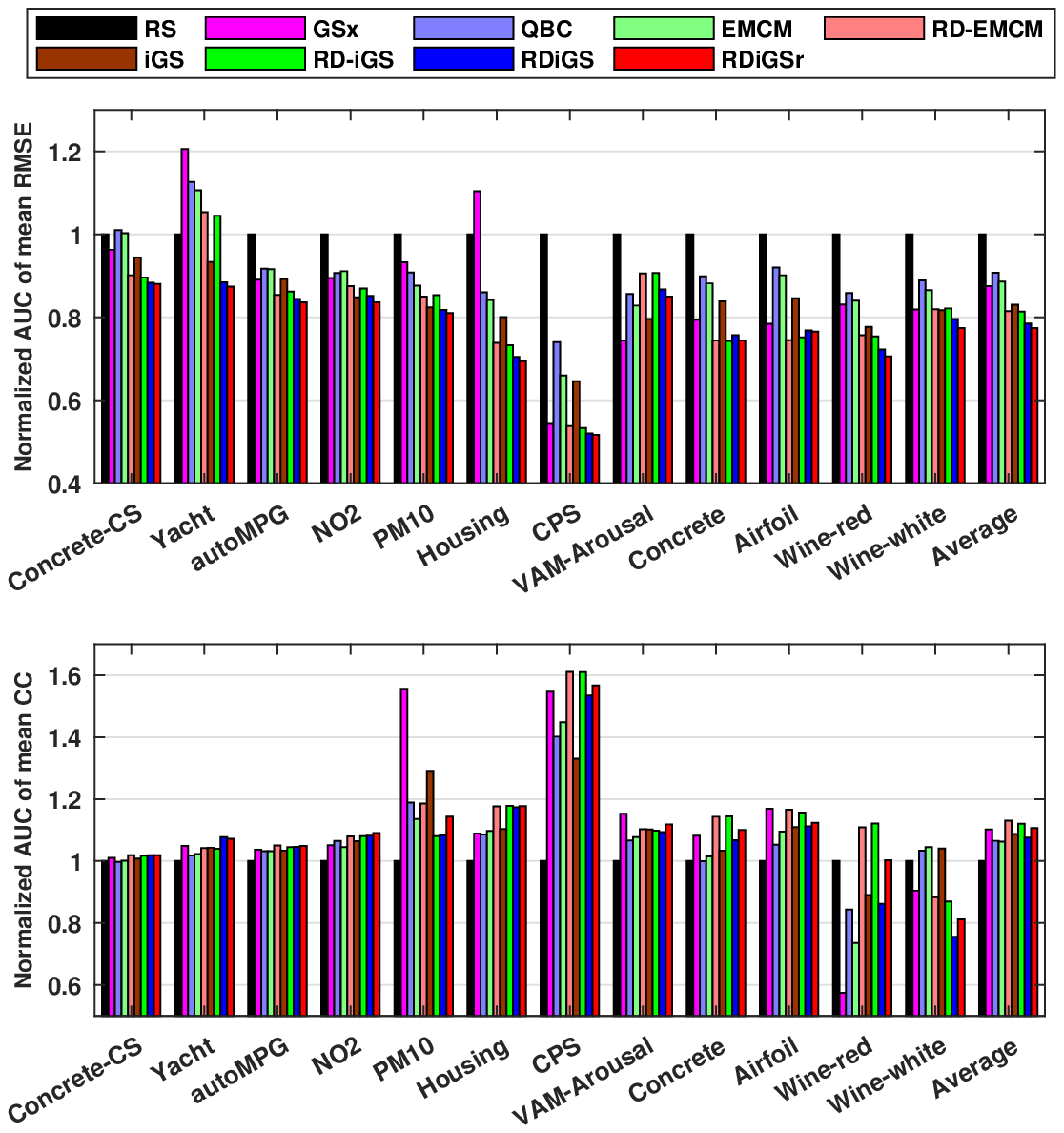}
\caption{Normalized AUCs of the mean RMSEs and the mean CCs on the 12 datasets. Ridge regression ($r=0.001$) was used as the regression model.} \label{fig:bar_9algs_ridge0001}
\end{figure}

\begin{figure}[!h]\centering
\includegraphics[width=\linewidth,clip]{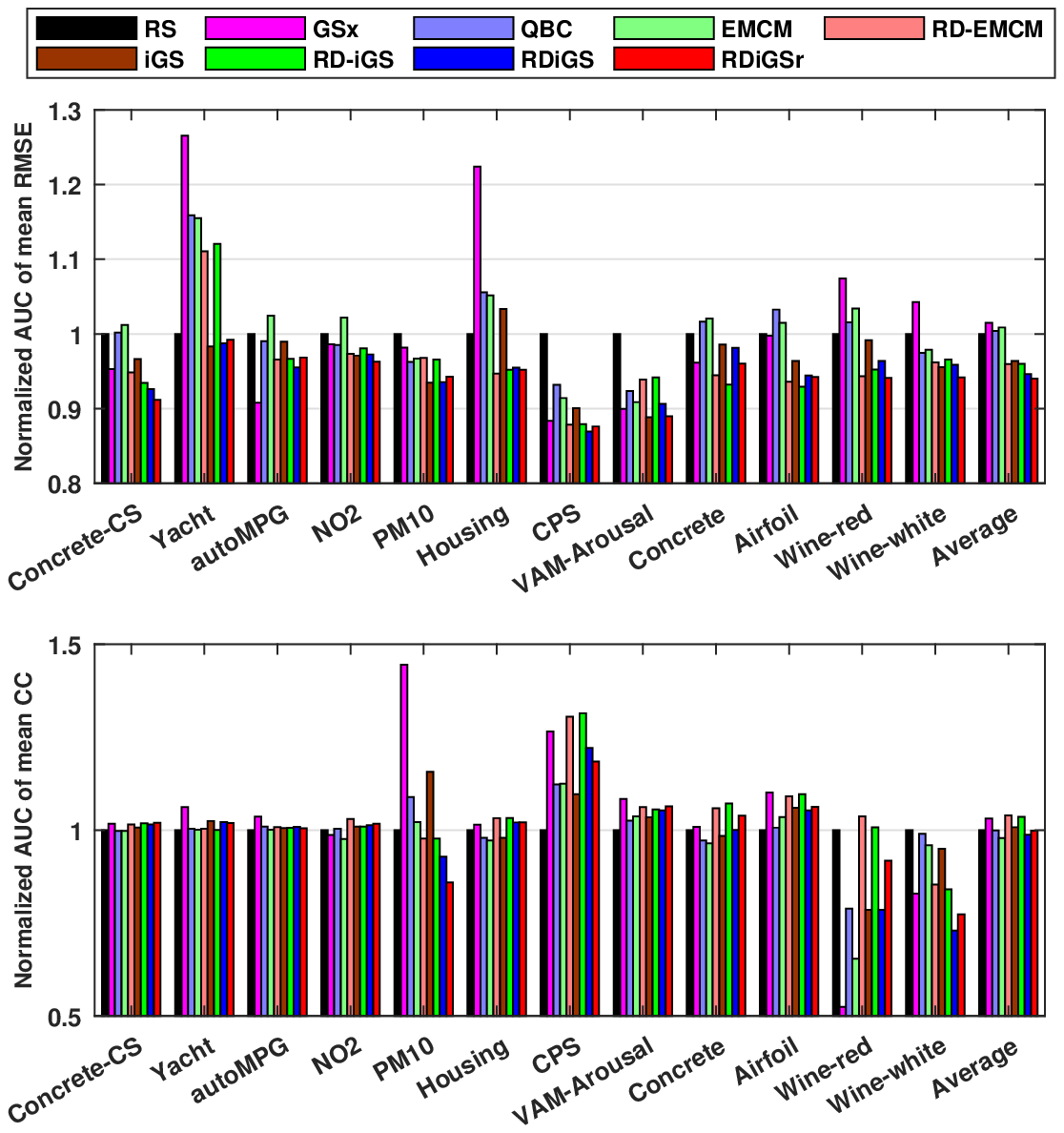}
\caption{Normalized AUCs of the mean RMSEs and the mean CCs on the 12 datasets. Ridge regression ($r=0.1$) was used as the regression model.} \label{fig:bar_9algs_ridge01}
\end{figure}

To determine whether our proposed approaches always perform better with different regularization coefficient, we repeated the experiments for different ridge regression models with different L2 regularization coefficient $r=0.001$ and $r=0.1$. The AUCs of the eight ALR approaches on the 12 datasets are shown in Figs.~\ref{fig:bar_9algs_ridge0001} and \ref{fig:bar_9algs_ridge01}, for $r=0.001$ and $r=0.1$, respectively. On average, RDiGSr and RDiGS still had smaller RMSE than RD-EMCM and iGS, and RDiGSr still had smaller RMSE and larger CC than RDiGS, for both $r=0.001$ and $r=0.1$. These results confirm again that RDiGSr was the best ALR approach among the eight.

\section{Conclusions} \label{sect:conclusions}

ALR is a frequently used machine learning approach for reducing the data labeling effort in regression problems. This paper considers pool-based sequential ALR, where a pool of unlabeled samples is given; we need to iteratively select some samples to label, from which a regression model can be trained to label other samples. Three essential criteria have to be considered in ALR: informativeness, representativeness, and diversity. We decompose ALR into two processes (initialization process and iteration process) and study how to incorporate these three criteria in each process separately. We proposed three new ALR approaches (RD-iGS, RDiGS and RDiGSr), corresponding to three different strategies for integrating the three criteria. RDiGS and RDiGSr performed much better than the state-of-the-art ALR approaches, and on average RDiGSr achieved the best performance.



\end{document}